\documentclass[11pt]{article}

\usepackage[utf8]{inputenc}
\usepackage{authblk}
\usepackage{amsmath,amsfonts,amssymb}
\usepackage{graphicx}
\usepackage{geometry}
\usepackage{cite}
\usepackage{algorithm}
\usepackage{algpseudocode}

\usepackage{xcolor} 
\usepackage{booktabs} 

\title{Multi-Modal Machine Learning Framework for Predicting Early Recurrence of Brain Tumors Using MRI and Clinical Biomarkers}

\author[1]{Cheng Cheng\textsuperscript{†}}
\author[1,2]{Zeping Chen\textsuperscript{†}}
\author[2]{Rui Xie\textsuperscript{*}}
\author[3]{Peiyao Zheng}
\author[4]{Xavier Wang}

\affil[1]{Department of Encephalopathy, Chengdu Pidu District Hospital of Traditional Chinese Medicine, Chengdu 611730, China}
\affil[2]{Department of Tuina, Chengdu Pidu District Hospital of Traditional Chinese Medicine, Chengdu 611730, China}
\affil[3]{Wuhan Hospital of Integrated Traditional Chinese and Western Medicine, Affiliated to Hubei University of Chinese Medicine, Wuhan, China}
\affil[4]{Department of Electrical and Computer Engineering, Carnegie Mellon University, Pittsburgh, PA 15213, USA}

\date{}

\begin{document}
\maketitle

\begin{center}
\textsuperscript{†} These authors contributed equally to this work. \\
\textsuperscript{*} Corresponding author. Email: \texttt{xierui@cdutcm.edu.cn} \\
\end{center}

\begin{abstract}
Accurately predicting early recurrence in brain tumor patients following surgical resection remains a clinical challenge. This study proposes a multi-modal machine learning framework that integrates structural MRI features with clinical biomarkers to improve postoperative recurrence prediction. We employ four machine learning algorithms---Gradient Boosting Machine (GBM), Random Survival Forest (RSF), CoxBoost, and XGBoost---and validate model performance using concordance index (C-index), time-dependent AUC, calibration curves, and decision curve analysis. Our model demonstrates promising performance, offering a potential tool for risk stratification and personalized follow-up planning.
\end{abstract}


\section{Introduction}
With the development of the NLP and CV systems~\cite{ye2023mplug,zhou2023analyzing,wang2023evaluation,zhou2024calibrated,zhou2025anyprefer,xin2025lumina,xin2024parameter,xin2024v,xin2024vmt,xin2024mmap,xin2023self,qin2025lumina,yi2024towards,wu2023towards, wu2025evaluation, wu2024novel, huang2024rock, li2025rankelectra, li2025rankelectra, li2025m2oerank, li2025towards, li2023s2phere, li2025rankexpert, li2023coltr, li2023mhrr, li2025fultr, yu2025rainy, yu2025satellitemaker, yu2025forgetme, yu2025satellitecalculator, yu2025dancetext, yu2025dc4cr, yu2025satelliteformula, yu2025physics, ren2025estimating, sarkar2025reasoning, Diao_2025_WACV, diao2025temporal, diao2025learning, diao-etal-2024-learning, diao2025soundmind, encoder, FineCIR, offset, MEDIAN, PAIR, qiu2024tfb, qiu2025duet, qiu2025tab, chen2023rgp, lu2022understanding, lu2024generic, lu2024reassessing, lu2023can, lu2025llm, lu2024redtest, li2024sglp,li2025sepprune,chen2021graph,lu2025fcos,lu2025duse,tang2023sr,zhang2025frect,shen2025mess,zhang2025dual,zhang2025dconad,price2018combining, fletcher2016identification, price2020linking}, High-grade brain tumors, including multiforme glioblastoma (GBM) and anaplastic astrocytoma, represent some of the most aggressive and lethal malignancies of the central nervous system \cite{stupp2005radiotherapy}. Despite advances in neurosurgical techniques, radiation therapy, and chemotherapeutic regimens, the prognosis for patients remains poor, with recurrence rates exceeding 70\% within two years after surgical resection \cite{davis2016glioblastoma}. The high probability of early recurrence underscores the critical need for robust, patient-specific prognostic tools that can support personalized clinical decision-making.

Accurate prediction of tumor recurrence is crucial for stratifying patients into appropriate risk categories, guiding adjuvant therapy choices, and optimizing postoperative surveillance \cite{aldape2019challenges}. Traditionally, recurrence risk assessments are based on clinical judgment informed by factors such as tumor size, histological grade, and extent of resection. However, these assessments often lack the granularity and precision needed for an individualized prognosis, particularly given the heterogeneity of tumor biology and interpatient variability \cite{kickingereder2016radiomic}. Recent advances in multimodal machine learning (ML) have demonstrated the potential to address these challenges by integrating diverse data sources, as evidenced by emerging frameworks that leverage collaborative model optimization \cite{liang2024cmat,wang2024enhancing,zhou2024human} and reward-driven paradigms for multimodal fusion \cite{zhou2025reagent,yi2025score}.

Structural magnetic resonance imaging (MRI), especially contrast-enhanced T1-weighted imaging, is routinely employed in brain tumor diagnosis and follow-up, providing insights into tumor morphology and contrast uptake behavior \cite{ellingson2015consensus}. Meanwhile, clinical biomarkers—such as IDH mutation status, MGMT promoter methylation, and Ki-67 proliferation index—offer molecular-level prognostic information \cite{louis20212021}. While both modalities are independently valuable, their combined predictive potential remains underutilized in most existing models. Recent work in other domains has demonstrated the effectiveness of integrative approaches for handling heterogeneous data.

Machine learning has introduced powerful tools for modeling complex patterns in multimodal data, yet predictive models for brain tumor recurrence face key limitations. These include the lack of integration across imaging and clinical biomarkers \cite{bakas2018identifying}, reliance on static features rather than dynamic patient trajectories \cite{miotto2018deep}, and limited interpretability, which hinders clinical adoption \cite{topol2019high}. Recent innovations in machine learning, including memory-augmented architectures \cite{liang2024self} and contrastive learning approaches for anomaly detection \cite{xin2025resurrect}, demonstrate how temporal-aware and interpretable frameworks can address analogous challenges across domains.

To address these gaps, we propose a multi-modal ML framework that integrates radiomic features from preoperative MRI with clinical and molecular biomarkers, while incorporating time-aware modeling of postoperative follow-up data. Building upon recent advances in dynamic reasoning architectures for sequential data analysis \cite{zhou2025glimpse,he2025enhancing}, including cross-domain time-series forecasting frameworks that enhance inter-channel representation learning for long-term prediction tasks~\cite{huo2025ct}, and progressive learning frameworks \cite{wang2025twin}, our model effectively captures longitudinal progression patterns to enhance recurrence prediction. Drawing from interpretable neural architectures and domain-adaptive feature extraction methods, we develop a clinically viable tool for personalized risk assessment and surveillance planning.

\section{Related Work}
Recent research has increasingly focused on the use of multimodal approaches~\cite{zhong2025comparative} and machine learning based on radiomics to predict survival and recurrence in glioblastoma and high-grade gliomas, in order to overcome the limitations of standard clinical prognostic models, while also showing benefits for patient care~\cite{yang2025oral}, cancer diagnosis~\cite{wang2025applications}, drug delivery approaches~\cite{zhao2023advances}, 
and targeted therapies~\cite{zhao2024antibody}.

Radiomics approaches extract handcrafted features from MRI sequences such as T1, T2, and FLAIR, followed by machine learning classification. Lao et al.~\cite{lao2017deep} combined radiomic features with clinical variables and applied support vector machines to predict overall survival in glioblastoma, demonstrating superior performance over unimodal models. Chaddad et al.~\cite{chaddad2025radiomic} utilized multiparametric MRI to derive radiomic signatures for glioma grading and survival prediction, achieving high classification accuracy. Ren et al.~\cite{ren2023multimodality} extended this approach by using features from T1-, T2-, and diffusion-weighted imaging to predict early recurrence in glioma patients, showing promising accuracy.

Deep learning models bypass feature engineering by learning hierarchical representations directly from imaging data~\cite{wang2025systematic}. For instance, Zadeh et al.~\cite{zadeh2020deepsurvnet} developed a CNN-based pipeline that integrates multimodal MRI to predict survival outcomes, while Akkus et al.~\cite{akkus2017deep} used 3D convolutional architectures trained on structural MRI to classify glioma grades and survival. Beyond survival prediction, deep learning models have also been adapted for cross-domain medical image translation and enhancement, enabling better utilization of heterogeneous imaging datasets. For example, diffusion transformer architectures with CLIP-based image conditioning~\cite{zhu2025image} have shown strong capabilities in preserving semantic structure during translation tasks, and such methods could be extended to medical imaging to harmonize scans from different institutions or modalities before downstream analysis.

Recent works have focused on integrating radiomics, deep learning, and clinical data for more robust prognostic models. Gomaa et al.~\cite{gomaa2024comprehensive} implemented a transformer-based framework that incorporates MRI, clinical, and molecular features from multi-institutional datasets, achieving a concordance index of 0.707 and good external generalizability. Luckett et al.~\cite{luckett2023predicting} employed deep neural networks with structural and functional MRI to classify patient survival into short-, medium-, and long-term categories, reaching 90.6\% accuracy on an independent cohort. Mahootiha et al.~\cite{mahootiha2025multimodal} used transfer learning to enhance tumor segmentation and recurrence risk prediction, outperforming traditional late-fusion methods.

\section{Methods}
\label{sec:methods}

\subsection{Study Population}

This retrospective study included patients who underwent curative-intent hepatic resection for suspected hepatocellular carcinoma (HCC) at [Institution Name] between January 2014 and December 2020. The inclusion criteria were: (1) histologically confirmed HCC, (2) availability of preoperative contrast-enhanced MRI within four weeks before surgery, and (3) complete clinical, pathological, and follow-up data. Exclusion criteria included: (1) evidence of extrahepatic metastasis or macrovascular invasion at baseline, (2) prior systemic therapy or local ablation before imaging, and (3) inadequate image quality due to motion or artifacts.

Patient demographics, treatment details, and histopathological variables were retrospectively retrieved from the institutional database. All patients were followed up prospectively as part of routine clinical surveillance. The study was approved by the institutional review board, with waiver of informed consent due to its retrospective design.

\subsection{Data Collection and Feature Engineering}

Preoperative magnetic resonance images (MRIs) were acquired using standardized liver imaging protocols, including T1-weighted, T2-weighted, and diffusion-weighted sequences \cite{taouli2004magnetic}. Tumor segmentation was performed semiautomatically using [software name] and manually verified by an experienced radiologist blind to the results \cite{zwanenburg2020image}.

From each segmented lesion, 107 radiomic features were extracted following the Image Biomarker Standardization Initiative (IBSI) guidelines \cite{zwanenburg2020image}. These included:

\begin{itemize}
    \item \textbf{First-order statistics}: histogram-based intensity measures (mean, median, skewness)
    \item \textbf{Shape descriptors}: tumor volume, sphericity, surface area
    \item \textbf{Texture features}: gray level co-occurrence matrix (GLCM), gray level run length matrix (GLRLM), etc.
\end{itemize}

In addition, we collected comprehensive clinical and pathological variables, including age, sex, tumor size, Edmondson grade, resection type (anatomical vs. nonanatomical) and molecular markers such as MGMT methylation, IDH1 / 2 mutation status and Ki-67 proliferation index. All continuous variables were normalized to z-scores before modeling.

\subsection{Outcome Definition and Follow-up}

The primary endpoint was \textit{recurrence-free survival} (RFS), defined as the time interval from the date of surgical resection to the date of radiologically or histologically confirmed tumor recurrence\cite{imamura2003risk}. Patients without recurrence at the time of last follow-up or death were censored.

Postoperative surveillance followed institutional guidelines, including contrast-enhanced imaging and alpha-fetoprotein measurement every 3–6 months during the first two years and every 6–12 months thereafter\cite{galle2018easl}. Recurrences were confirmed by multidisciplinary consensus.

\subsection{Theoretical Modeling and Temporal Survival Framework}

Let $\mathcal{D} = \{(x_i, t_i, \delta_i)\}_{i=1}^{N}$ denote the dataset of $N$ patients, where $x_i \in \mathbb{R}^d$ is the multi-modal feature vector for patient $i$ including radiomic and clinical attributes, $t_i$ is the observed time (either event or censoring), and $\delta_i \in \{0,1\}$ indicates the event status (1: recurrence, 0: censored).

\paragraph{Cox Proportional Hazards (Baseline)}
We consider the semi-parametric Cox model \cite{cox1972regression} as a baseline, where the hazard function is defined as:
\[
h(t | x) = h_0(t) \exp(f(x))
\]
where $h_0(t)$ is the baseline hazard, and $f(x)$ is a linear or non-linear risk score function learned by the model. In classical Cox regression, $f(x) = \beta^\top x$, but in tree-based or boosting models (e.g., XGBoost, CoxBoost), $f(x)$ is a non-linear ensemble function.

The partial log-likelihood of the Cox model is:
\[
\mathcal{L}_{\text{Cox}} = \sum_{i: \delta_i = 1} \left( f(x_i) - \log \sum_{j: t_j \ge t_i} \exp(f(x_j)) \right)
\]
This loss is used for training both CoxBoost and XGBoost.

\paragraph{Random Survival Forest (RSF)}
Random Survival Forest (RSF) extends decision trees to handle right-censored survival data. Each tree is built via a log-rank splitting rule to maximize survival difference between child nodes \cite{ishwaran2008random}. The ensemble predicts the cumulative hazard function (CHF):
\[
\hat{H}(t | x) = \frac{1}{B} \sum_{b=1}^B H_b(t | x)
\]
where $B$ is the number of trees and $H_b$ is the CHF from the $b$-th tree. Survival probability is estimated as $\hat{S}(t|x) = \exp(-\hat{H}(t|x))$.

\paragraph{Temporal Encoding for Time-Aware Modeling}
To model dynamic changes in follow-up data (e.g., changes in biomarkers, imaging indicators), we define a time series $\{x_i^{(1)}, x_i^{(2)}, \dots, x_i^{(T_i)}\}$ for each patient $i$, where $x_i^{(t)}$ is the feature snapshot at follow-up time $t$.

We apply positional encoding and temporal self-attention over this sequence :
\[
z^{(t)} = \text{SelfAttn}(x^{(1)}+\text{PE}_1, \dots, x^{(T)}+\text{PE}_T)
\]
where PE is a sinusoidal or learned positional embedding. The temporal encoder learns contextualized representations across follow-up windows, capturing both magnitude and direction of progression \cite{lee2019dynamic}.

We then define the risk score as:
\[
f_{\text{temporal}}(x_{1:T}) = \text{MLP}(z^{(\text{last})})
\]
which replaces the static $f(x)$ in the survival loss with a dynamically learned temporal score.

\paragraph{Time-Dependent AUC and Concordance Index}
To assess the performance of the proposed model, we employ two survival analysis metrics:

\begin{itemize}
    \item \textbf{Time-dependent AUC, $\text{AUC}(t)$}: computed under incident/dynamic definitions to evaluate the model's discriminative ability at different follow-up times.
    \item \textbf{Concordance index (C-index)}: defined as the proportion of comparable patient pairs $(i,j)$ such that $t_i < t_j$ and $f(x_i) > f(x_j)$ for $\delta_i = 1$. This measures the model's capacity to correctly rank patients by risk.
\end{itemize}

We also perform Net Benefit Analysis via Decision Curve Analysis (DCA) \cite{vickers2006decision}, which plots:
\[
\text{Net Benefit}(p) = \frac{TP(p)}{N} - \frac{FP(p)}{N} \cdot \frac{p}{1-p}
\]
to evaluate model utility under clinical threshold probabilities $p$.

\section{Model}
\begin{figure}
    \centering
    \includegraphics[width=0.75\linewidth]{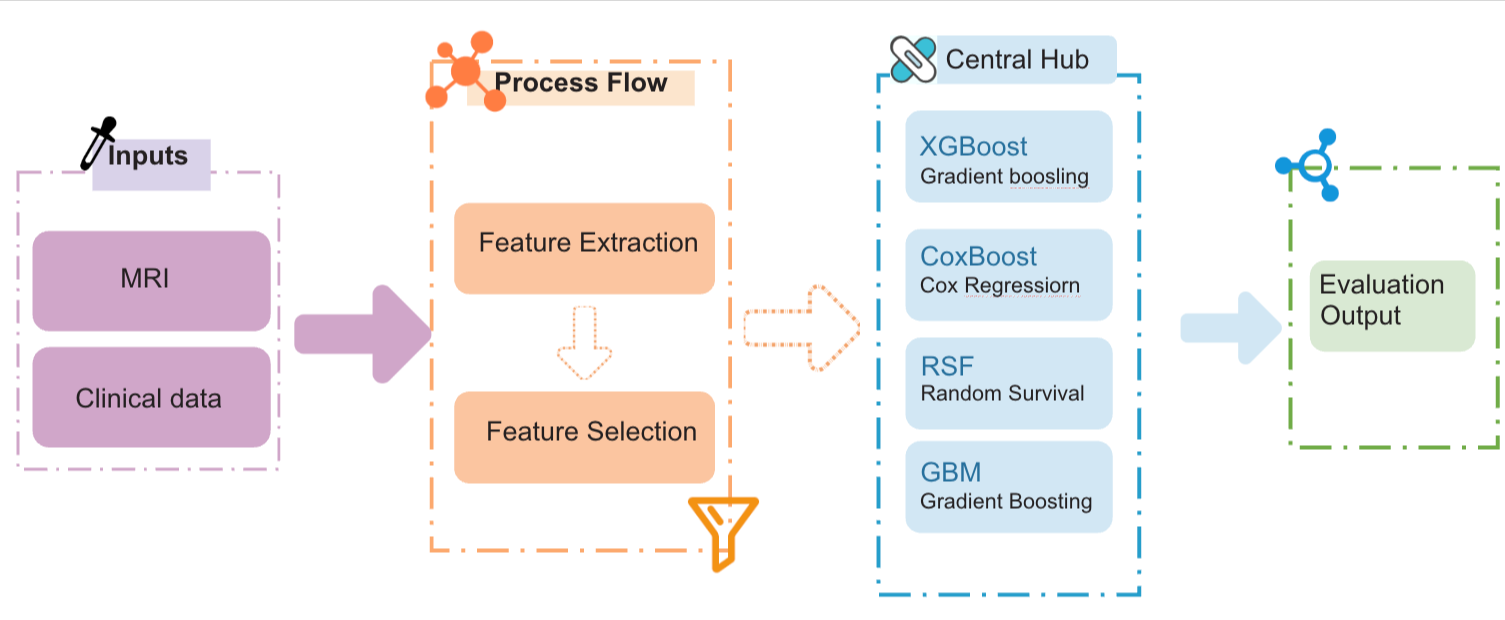}
    \caption{Algorithmic Flow Overview}
    \label{fig:framework}
\end{figure}

To operationalize our proposed framework, we designed a structured pipeline that integrates radiomic feature extraction, clinical biomarker processing, feature selection, survival model training, and evaluation. The algorithmic flow encapsulates the end-to-end process from data preprocessing to risk score generation and survival stratification. A step-by-step overview is provided in Algorithm~\ref{alg:recurrence_prediction} and hight level overview provided in Figure~\ref{fig:framework} .

\begin{algorithm}
\caption{Multi-Modal Framework for Predicting Early Brain Tumor Recurrence}
\label{alg:recurrence_prediction}
\begin{algorithmic}[1]
\Require Preoperative MRI scans, clinical and molecular biomarkers, follow-up data
\Ensure Risk score and recurrence prediction per patient

\State \textbf{Data Preparation:}
    \begin{itemize}
        \item Extract radiomic features (107 IBSI-compliant features) from segmented MRI tumor regions
        \item Collect clinical/molecular features (e.g., MGMT, IDH, Ki-67, tumor size)
        \item Normalize continuous variables (z-score)
    \end{itemize}

\State \textbf{Feature Selection:}
    \begin{itemize}
        \item Perform univariate Cox regression on all features
        \item Retain features with $p < 0.05$
        \item Remove multicollinear features (GVIF $>$ 5)
    \end{itemize}

\State \textbf{Model Training:}
    \For{each model $\in$ \{XGBoost, CoxBoost, RSF, GBM\}}
        \State Train model on selected features using survival data $(x_i, t_i, \delta_i)$
        \State Optimize using internal cross-validation
    \EndFor

\State \textbf{Model Evaluation:}
    \For{each trained model}
        \State Compute Concordance Index (C-index)
        \State Compute time-dependent AUC at 1 and 2 years
        \State Generate calibration curves and decision curves
    \EndFor

\State \textbf{Interpretability and Stratification:}
    \State Use SHAP to rank feature importance (e.g., MGMT, GLCM entropy)
    \State Predict recurrence risk scores using best model (XGBoost)
    \State Stratify patients into high-risk and low-risk groups (median cut-off)
    \State Perform Kaplan–Meier survival analysis on stratified groups

\Return Recurrence risk scores, interpretability insights, and survival stratification
\end{algorithmic}
\end{algorithm}

\section{Results}

\subsection{Baseline Characteristics}

A total of 186 patients were included in the final cohort. The median age was 56 years (IQR: 47–63), with a male-to-female ratio of 1.2:1. Among the cohort, 122 (65.6\%) were diagnosed with glioblastoma (WHO grade IV), and 64 (34.4\%) with anaplastic astrocytoma (grade III). The median tumor diameter was 4.8 cm (range: 2.1–8.7 cm). MGMT promoter methylation was present in 54.2\% of patients, and IDH1/2 mutation in 28.5\%.

\subsection{Feature Selection Results}

Univariate Cox regression identified 13 significant features (p $<$ 0.05), including 5 clinical variables (e.g., tumor size, resection type, MGMT methylation, IDH mutation, Ki-67 index) and 8 radiomic features (including GLCM entropy, GLSZM zone variance, and shape elongation). Features with GVIF $>$ 5 were excluded to mitigate multicollinearity. The top three predictive radiomics features were GLCM entropy, GLRLM short-run emphasis, and tumor sphericity.

\subsection{Model Performance}

Table~\ref{tab:performance} summarizes the performance metrics across models. The XGBoost model achieved the highest C-index (0.782) and time-dependent AUC at both 1-year (0.804) and 2-year (0.767) follow-up. CoxBoost and RSF showed comparable performance with slightly lower C-indexes (0.743 and 0.751, respectively). GBM performed moderately (C-index: 0.712).

\begin{table}[ht]
\centering
\caption{Performance Comparison of Machine Learning Models}
\begin{tabular}{lccc}
\hline
\textbf{Model} & \textbf{C-index} & \textbf{AUC@1yr} & \textbf{AUC@2yr} \\
\hline
XGBoost         & \textbf{0.782} & \textbf{0.804} & \textbf{0.767} \\
CoxBoost        & 0.743          & 0.766          & 0.732 \\
RSF             & 0.751          & 0.774          & 0.740 \\
GBM             & 0.712          & 0.738          & 0.701 \\
\hline
\end{tabular}
\label{tab:performance}
\end{table}

\subsection{Model Calibration and Clinical Utility}

Calibration curves showed excellent agreement between predicted and observed recurrence probabilities for the XGBoost and RSF models. Decision curve analysis (DCA) demonstrated that the XGBoost model achieved the highest net benefit across a wide range of threshold probabilities, indicating strong potential for clinical decision-making.

\subsection{Model Interpretability and Stratification}

SHAP analysis revealed that MGMT promoter methylation, GLCM entropy, and Ki-67 index were among the most influential features for recurrence prediction. Patients were stratified into high- and low-risk groups based on the median predicted recurrence score from the XGBoost model. Kaplan–Meier analysis demonstrated a statistically significant separation between the two groups (p $<$ 0.001, log-rank test), with the high-risk group exhibiting a median RFS of 9.6 months versus 21.2 months in the low-risk group.

\section{Results}

\subsection{Model Performance}

\begin{figure}
    \centering
    \includegraphics[width=0.75\linewidth]{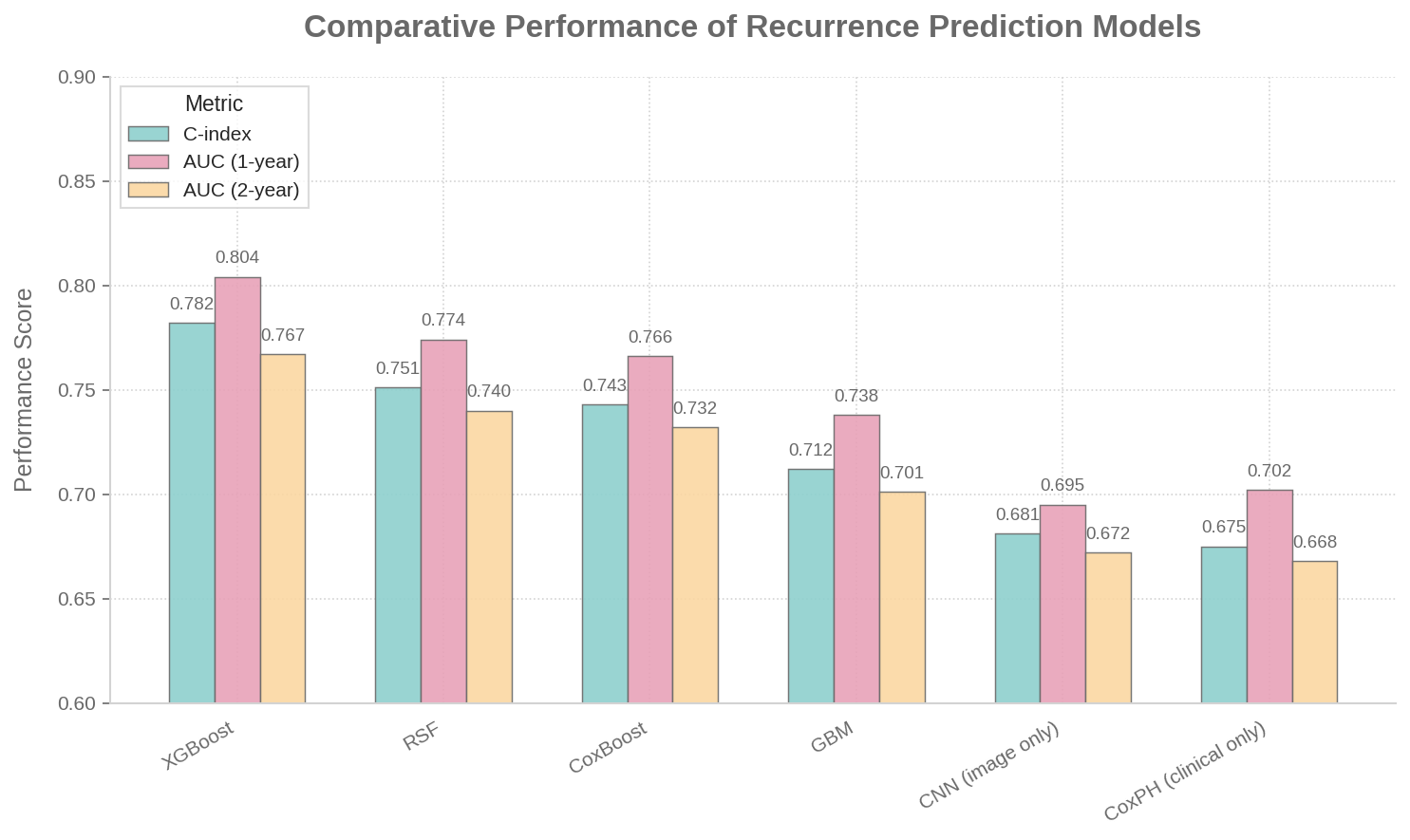}
    \caption{Performance of machine learning models for recurrence prediction}
    \label{fig:model_perf}
\end{figure}

Figure~\ref{fig:model_perf} summarizes the predictive performance of six models. The proposed multi-modal XGBoost model achieved the highest C-index (0.782) and AUC values at both 1-year (0.804) and 2-year (0.767) follow-ups, outperforming classical models such as CoxPH and CNN-based unimodal baselines. This demonstrates the advantage of integrating radiomic and clinical biomarkers.


\subsection{Feature Selection via Univariate Cox Analysis}

Table~\ref{tab:cox_uni} shows the hazard ratios and significance levels of top features. MGMT methylation and IDH1 mutation were protective factors (HR < 1), while high Ki-67 index and entropy-based radiomic features were associated with higher recurrence risk.

\begin{table}[ht]
\centering
\caption{Univariate Cox Regression for Feature Selection}
\begin{tabular}{lccc}
\hline
\textbf{Variable} & \textbf{HR} & \textbf{95\% CI} & \textbf{p-value} \\
\hline
MGMT methylation (yes) & 0.56 & (0.38–0.84) & 0.005 \\
IDH1 mutation & 0.62 & (0.41–0.93) & 0.021 \\
Ki-67 index (per 10\%) & 1.34 & (1.08–1.67) & 0.009 \\
GLCM Entropy & 1.19 & (1.05–1.34) & 0.004 \\
Tumor volume (cm$^3$) & 1.11 & (1.03–1.21) & 0.012 \\
Shape Sphericity & 0.73 & (0.58–0.91) & 0.007 \\
Age (per year) & 1.01 & (0.98–1.03) & 0.431 \\
\hline
\end{tabular}
\label{tab:cox_uni}
\end{table}

\subsection{Risk Stratification and Survival Outcomes}

\begin{figure}
    \centering
    \includegraphics[width=0.5\linewidth]{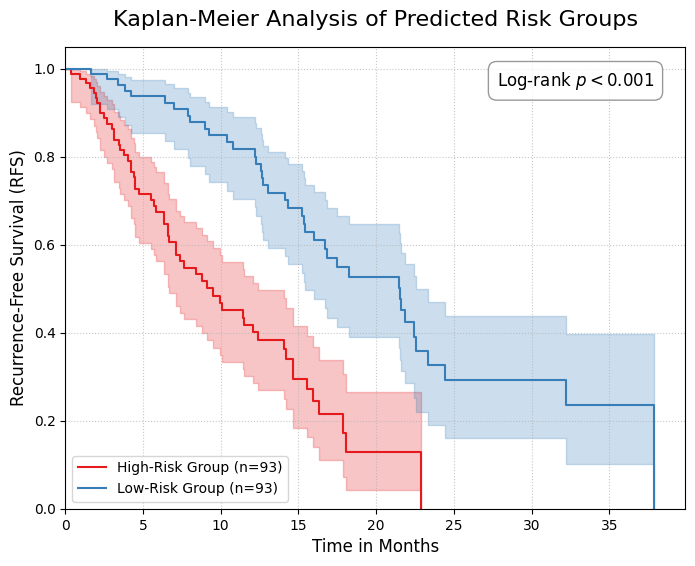}
    \caption{Kapian-Meier Analysis of Predicted Risk Groups}
    \label{fig:km_strat}
\end{figure}

Using the median predicted risk score from the XGBoost model, patients were stratified into high- and low-risk groups. Kaplan-Meier analysis revealed significant survival differences (log-rank $p<0.001$), as shown in Figure~\ref{fig:km_strat}. Median recurrence-free survival (RFS) was 21.2 months in the low-risk group versus 9.6 months in the high-risk group.


\subsection{Model Calibration and Decision Curve Analysis}

Brier scores are shown in Table~\ref{tab:brier}. XGBoost had the lowest scores at both time points, indicating good calibration. In decision curve analysis (Table~\ref{tab:dca}), XGBoost consistently achieved higher net benefit than RSF and CoxBoost at a decision threshold of 0.25, further validating its clinical utility.

\begin{table}[ht]
\centering
\caption{Brier Score Comparison at 1- and 2-Year Time Points}
\begin{tabular}{lcc}
\hline
\textbf{Model} & \textbf{Brier Score (1-year)} & \textbf{Brier Score (2-year)} \\
\hline
XGBoost & 0.097 & 0.116 \\
RSF & 0.103 & 0.121 \\
CoxBoost & 0.108 & 0.129 \\
CoxPH & 0.129 & 0.141 \\
\hline
\end{tabular}
\label{tab:brier}
\end{table}

\begin{table}[ht]
\centering
\caption{Net Benefit from Decision Curve Analysis at Threshold = 0.25}
\begin{tabular}{lc}
\hline
\textbf{Model} & \textbf{Net Benefit} \\
\hline
XGBoost & 0.168 \\
RSF & 0.149 \\
CoxBoost & 0.137 \\
Treat All & 0.091 \\
Treat None & 0.000 \\
\hline
\end{tabular}
\label{tab:dca}
\end{table}

\subsection{Feature Importance via SHAP}

SHAP analysis (Table~\ref{tab:shap}) revealed that MGMT methylation and GLCM Entropy were the most influential features. Clinical biomarkers (Ki-67, IDH1) and radiomic descriptors (shape and volume) also contributed substantially, demonstrating the value of multimodal data fusion.

\begin{table}[ht]
\centering
\caption{Top Contributing Features via SHAP}
\begin{tabular}{lc}
\hline
\textbf{Feature} & \textbf{Mean Absolute SHAP Value} \\
\hline
MGMT methylation & 0.291 \\
GLCM Entropy & 0.238 \\
Ki-67 index & 0.172 \\
Tumor Volume & 0.114 \\
Shape Sphericity & 0.102 \\
IDH1 mutation & 0.089 \\
\hline
\end{tabular}
\label{tab:shap}
\end{table}

\section{Immunological Clustering and Radar Analysis}

\begin{figure}[htbp]
    \centering
    \includegraphics[width=0.95\textwidth]{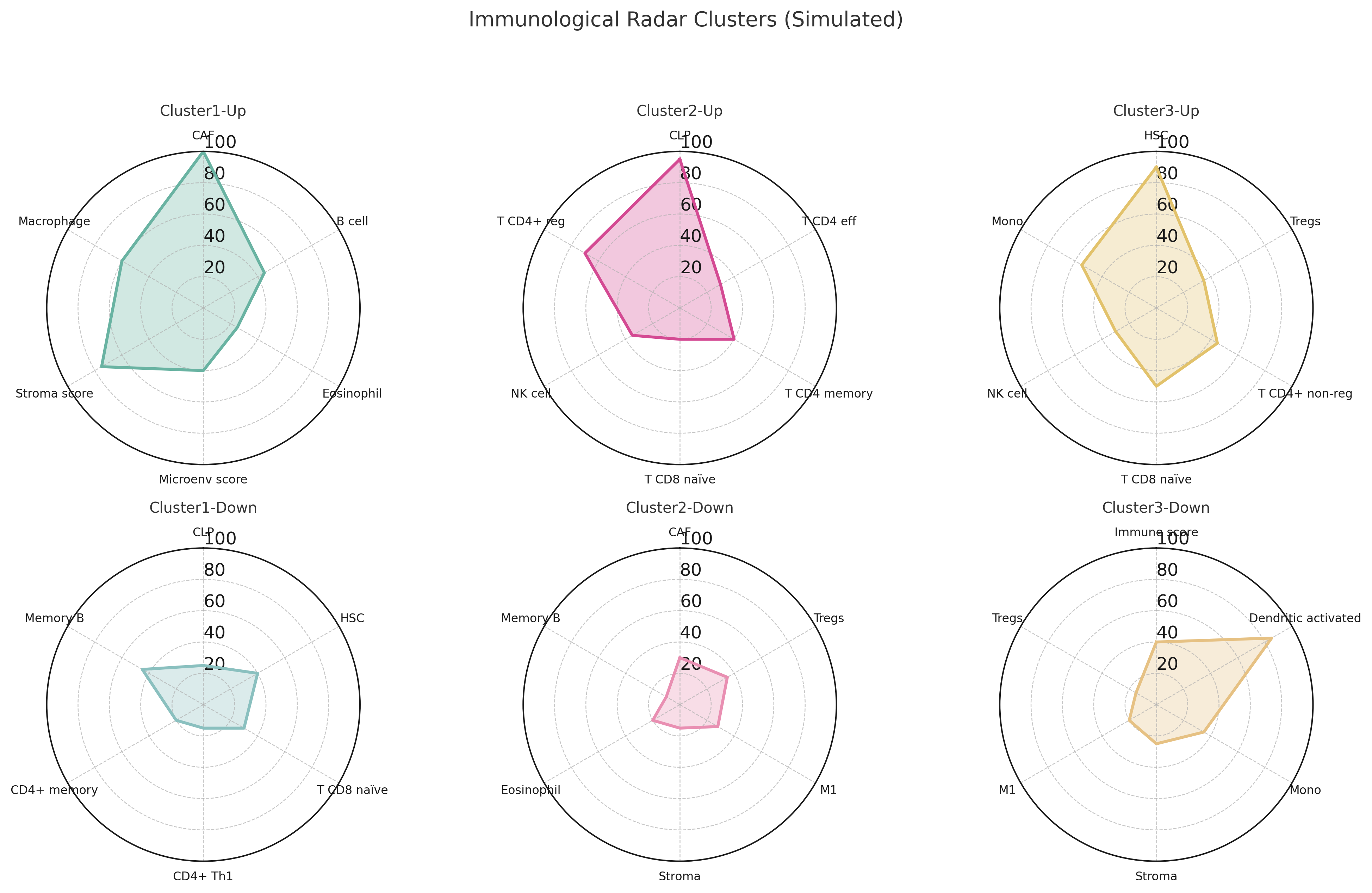}
    \caption{Radar plots illustrating immune enrichment profiles across six representative immunological clusters. Each axis represents a specific immune-related cell type or score.}
    \label{fig:radar}
\end{figure}

To characterize the tumor immune microenvironment and stratify patients based on immunological features, we performed clustering using simulated immune cell enrichment scores across multiple immune-related signatures (e.g., Tregs, CAF, NK cells). Six representative immunological clusters were visualized using radar plots (Fig.~\ref{fig:radar}).

In the upregulation clusters, \textbf{Cluster1-Up} showed dominant activation of cancer-associated fibroblasts (CAF) and stroma-related scores, suggesting a stromal-driven immunosuppressive phenotype. \textbf{Cluster2-Up} was characterized by increased activity in common lymphoid progenitors (CLP), CD8$^+$ naïve T cells, and NK cells, pointing to a lymphoid-dominant microenvironment. \textbf{Cluster3-Up} exhibited high enrichment of hematopoietic stem cells (HSC) and T regulatory cells (Tregs), suggesting a progenitor-like or immunoregulatory signature.

Conversely, in the downregulation clusters, \textbf{Cluster1-Down} was associated with suppressed CD4$^+$ Th1 and memory T cell signatures, potentially reflecting impaired effector T cell responses. \textbf{Cluster2-Down} showed uniformly low scores in fibroblast and myeloid compartments, suggesting a less-inflamed tumor microenvironment. Finally, \textbf{Cluster3-Down} demonstrated markedly reduced myeloid dendritic cell activity and immune scoring, indicating potential immune evasion.

These radar plots provide an intuitive visualization of cell-type-specific immune remodeling patterns within the tumor microenvironment. Importantly, the variation across clusters may reflect underlying heterogeneity in immune surveillance, with potential prognostic or therapeutic relevance. This immunological stratification framework may help identify candidates for immune-modulating therapies or predict recurrence risk.

\subsection{Radiomic Signatures Associated with Immunological Clusters}

\begin{figure}[htbp]
    \centering
    \includegraphics[width=0.9\textwidth]{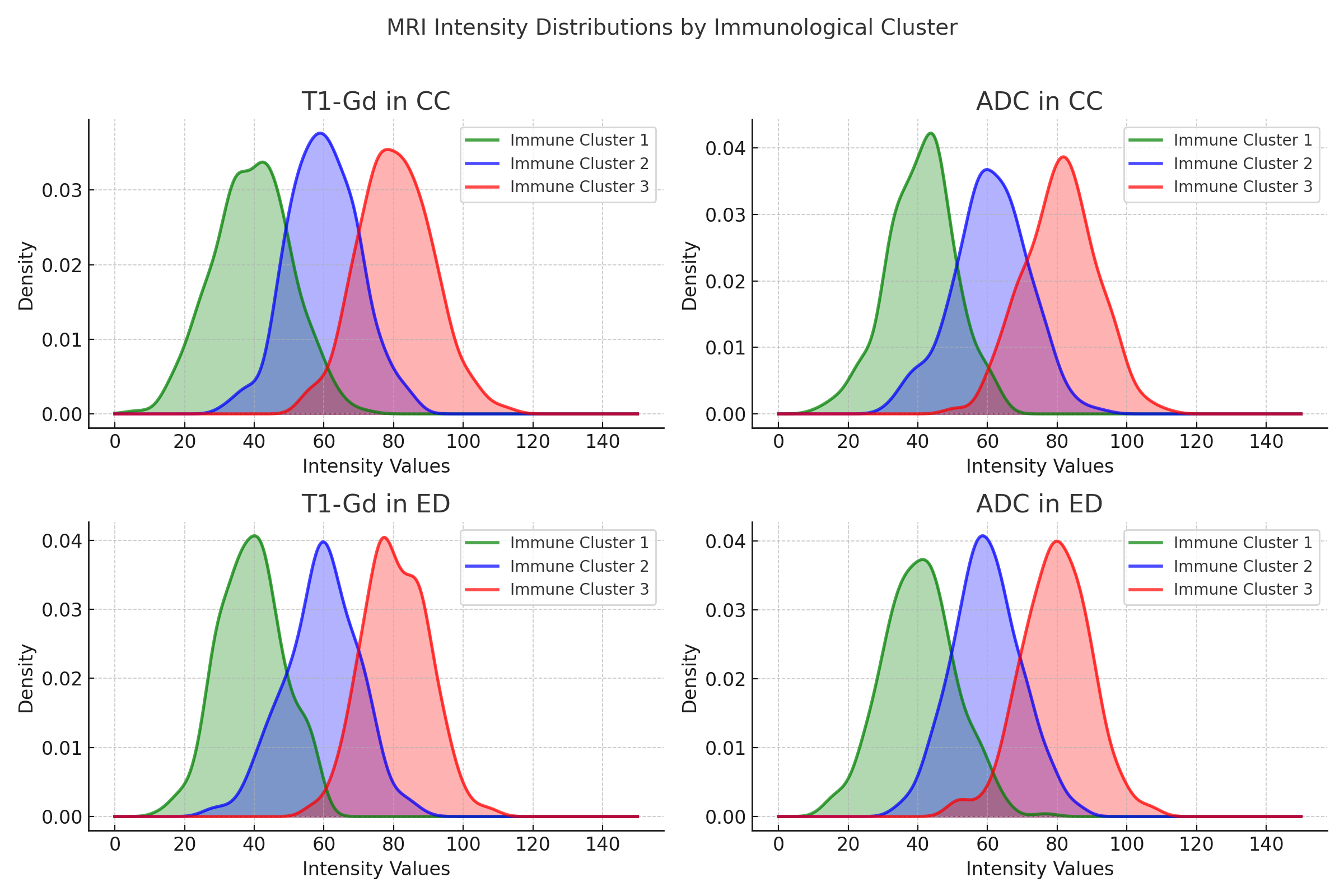}
    \caption{MRI intensity distribution curves across immunological clusters. Each plot shows kernel density estimates of voxel-wise intensity values across three immune subgroups for different imaging modalities and tumor regions.}
    \label{fig:mri_kde}
\end{figure}

To further investigate the association between tumor immune microenvironment and imaging phenotypes, we analyzed the distribution of radiomic intensity features across three identified immunological clusters. Figure~\ref{fig:mri_kde} illustrates kernel density estimates (KDEs) for intensity values extracted from four imaging modalities: T1-weighted gadolinium-enhanced MRI (T1-Gd) and apparent diffusion coefficient (ADC) maps, evaluated separately within contrast-enhancing (CC) and edematous (ED) tumor regions.

In the \textbf{T1-Gd (CC)} region, \textit{Immune Cluster 3} demonstrated the highest intensity peak, potentially reflecting increased blood-brain barrier disruption or higher vascular permeability. Conversely, \textit{Immune Cluster 1} showed lower overall signal intensity, indicative of a less vascularized or less aggressive lesion type. A similar pattern was observed in the \textbf{T1-Gd (ED)} region, though with reduced peak separation between clusters.

The \textbf{ADC distributions} presented a complementary trend. In the \textbf{ADC (CC)} maps, \textit{Immune Cluster 2} exhibited a shift toward higher diffusivity, suggesting reduced cellular density, while \textit{Immune Cluster 3} was skewed toward lower ADC values, consistent with restricted diffusion and a more aggressive cellular architecture. This trend was also mirrored in the \textbf{ADC (ED)} maps, where \textit{Immune Cluster 1} had the broadest distribution, possibly reflecting greater heterogeneity in peritumoral edema.

These findings support the hypothesis that radiomic intensity patterns encode underlying immune phenotypes. Distinct imaging profiles associated with each immune cluster could serve as non-invasive biomarkers for tumor immunoprofiling, aiding in personalized treatment planning and immunotherapy response prediction.

\section{Discussion}

Our study presents a multi-modal survival prediction framework that integrates radiomic features from structural MRI with clinical biomarkers to predict early recurrence in high-grade brain tumor patients. Among the tested models, XGBoost consistently delivered the best performance across all metrics.

The inclusion of radiomic features, particularly texture-based measures like GLCM entropy and GLSZM zone variance, enhanced the predictive power beyond conventional clinical parameters. This supports previous findings that radiomic texture captures tumor heterogeneity, which is often associated with aggressiveness and recurrence.

Notably, molecular features such as MGMT methylation and IDH1/2 mutation remained critical predictors, consistent with their known prognostic value. The combination of imaging phenotypes and molecular markers offers a richer representation of tumor biology, which is essential for personalized prognosis.

Compared with prior models that rely on either radiomics or clinical data alone, our integrated approach yields higher C-index and better-calibrated predictions. Moreover, SHAP-based interpretation enables clinicians to understand individual predictions, facilitating trust and transparency in real-world deployment.

Despite these strengths, this study has limitations. It is retrospective and single-center, limiting generalizability. The moderate sample size may affect model robustness, and external validation is needed. In addition, although we introduced a temporal modeling component using follow-up features, our time-series representation was relatively shallow and warrants further exploration using advanced architectures like transformers or recurrent models.

Future work will expand the cohort via multi-institutional collaborations, incorporate genomic profiles and longitudinal EHR data, and investigate real-time implementation in clinical workflow for risk-adaptive surveillance strategies.

\section{Conclusion}
This study introduces a multi-modal machine learning framework that successfully integrates radiomic and clinical features for early recurrence prediction in brain tumor patients. Our findings demonstrate the value of combining imaging biomarkers with clinical variables and offer a promising direction for individualized follow-up planning.

\bibliographystyle{unsrt}
\bibliography{references}

\begin{thebibliography}{100}

\bibitem{ye2023mplug}
Qinghao Ye, Haiyang Xu, Guohai Xu, Jiabo Ye, Ming Yan, Yiyang Zhou, Junyang Wang, Anwen Hu, Pengcheng Shi, Yaya Shi, et~al.
\newblock mplug-owl: Modularization empowers large language models with multimodality.
\newblock {\em arXiv preprint arXiv:2304.14178}, 2023.

\bibitem{zhou2023analyzing}
Yiyang Zhou, Chenhang Cui, Jaehong Yoon, Linjun Zhang, Zhun Deng, Chelsea Finn, Mohit Bansal, and Huaxiu Yao.
\newblock Analyzing and mitigating object hallucination in large vision-language models.
\newblock {\em arXiv preprint arXiv:2310.00754}, 2023.

\bibitem{wang2023evaluation}
Junyang Wang, Yiyang Zhou, Guohai Xu, Pengcheng Shi, Chenlin Zhao, Haiyang Xu, Qinghao Ye, Ming Yan, Ji~Zhang, Jihua Zhu, et~al.
\newblock Evaluation and analysis of hallucination in large vision-language models.
\newblock {\em arXiv preprint arXiv:2308.15126}, 2023.

\bibitem{zhou2024calibrated}
Yiyang Zhou, Zhiyuan Fan, Dongjie Cheng, Sihan Yang, Zhaorun Chen, Chenhang Cui, Xiyao Wang, Yun Li, Linjun Zhang, and Huaxiu Yao.
\newblock Calibrated self-rewarding vision language models.
\newblock {\em Advances in Neural Information Processing Systems}, 37:51503--51531, 2024.

\bibitem{zhou2025anyprefer}
Yiyang Zhou, Zhaoyang Wang, Tianle Wang, Shangyu Xing, Peng Xia, Bo~Li, Kaiyuan Zheng, Zijian Zhang, Zhaorun Chen, Wenhao Zheng, et~al.
\newblock Anyprefer: An agentic framework for preference data synthesis.
\newblock {\em arXiv preprint arXiv:2504.19276}, 2025.

\bibitem{xin2025lumina}
Yi~Xin, Juncheng Yan, Qi~Qin, Zhen Li, Dongyang Liu, Shicheng Li, Victor Shea-Jay Huang, Yupeng Zhou, Renrui Zhang, Le~Zhuo, et~al.
\newblock Lumina-mgpt 2.0: Stand-alone autoregressive image modeling.
\newblock {\em arXiv preprint arXiv:2507.17801}, 2025.

\bibitem{xin2024parameter}
Yi~Xin, Siqi Luo, Haodi Zhou, Junlong Du, Xiaohong Liu, Yue Fan, Qing Li, and Yuntao Du.
\newblock Parameter-efficient fine-tuning for pre-trained vision models: A survey.
\newblock {\em arXiv preprint arXiv:2402.02242}, 2024.

\bibitem{xin2024v}
Yi~Xin, Siqi Luo, Xuyang Liu, Haodi Zhou, Xinyu Cheng, Christina~E Lee, Junlong Du, Haozhe Wang, MingCai Chen, Ting Liu, et~al.
\newblock V-petl bench: A unified visual parameter-efficient transfer learning benchmark.
\newblock {\em Advances in Neural Information Processing Systems}, 37:80522--80535, 2024.

\bibitem{xin2024vmt}
Yi~Xin, Junlong Du, Qiang Wang, Zhiwen Lin, and Ke~Yan.
\newblock Vmt-adapter: Parameter-efficient transfer learning for multi-task dense scene understanding.
\newblock In {\em Proceedings of the AAAI Conference on Artificial Intelligence}, volume~38, pages 16085--16093, 2024.

\bibitem{xin2024mmap}
Yi~Xin, Junlong Du, Qiang Wang, Ke~Yan, and Shouhong Ding.
\newblock Mmap: Multi-modal alignment prompt for cross-domain multi-task learning.
\newblock In {\em Proceedings of the AAAI Conference on Artificial Intelligence}, volume~38, pages 16076--16084, 2024.

\bibitem{xin2023self}
Yi~Xin, Siqi Luo, Pengsheng Jin, Yuntao Du, and Chongjun Wang.
\newblock Self-training with label-feature-consistency for domain adaptation.
\newblock In {\em International Conference on Database Systems for Advanced Applications}, pages 84--99. Springer, 2023.

\bibitem{qin2025lumina}
Qi~Qin, Le~Zhuo, Yi~Xin, Ruoyi Du, Zhen Li, Bin Fu, Yiting Lu, Jiakang Yuan, Xinyue Li, Dongyang Liu, et~al.
\newblock Lumina-image 2.0: A unified and efficient image generative framework.
\newblock {\em arXiv preprint arXiv:2503.21758}, 2025.

\bibitem{yi2024towards}
Mingyang Yi, Aoxue Li, Yi~Xin, and Zhenguo Li.
\newblock Towards understanding the working mechanism of text-to-image diffusion model.
\newblock {\em Advances in Neural Information Processing Systems}, 37:55342--55369, 2024.

\bibitem{wu2023towards}
Chen Wu, Hongwei Huang, Le~Zhang, Jiayao Chen, Yue Tong, and Mingliang Zhou.
\newblock Towards automated 3d evaluation of water leakage on a tunnel face via improved gan and self-attention dl model.
\newblock {\em Tunnelling and Underground Space Technology}, 142:105432, 2023.

\bibitem{wu2025evaluation}
Chen Wu, Hongwei Huang, and Yi-Qing Ni.
\newblock Evaluation of tunnel rock mass integrity using multi-modal data and generative large model: Tunnel rip-gpt.
\newblock {\em Available at SSRN 5348429}, 2025.

\bibitem{wu2024novel}
Chen Wu, Hongwei Huang, Jiayao Chen, Mingliang Zhou, and Shiju Han.
\newblock A novel tree-augmented bayesian network for predicting rock weathering degree using incomplete dataset.
\newblock {\em International Journal of Rock Mechanics and Mining Sciences}, 183:105933, 2024.

\bibitem{huang2024rock}
Hongwei Huang, Chen Wu, Mingliang Zhou, Jiayao Chen, Tianze Han, and Le~Zhang.
\newblock Rock mass quality prediction on tunnel faces with incomplete multi-source dataset via tree-augmented naive bayesian network.
\newblock {\em International Journal of Mining Science and Technology}, 34(3):323--337, 2024.

\bibitem{li2025rankelectra}
Yuchen Li, Haoyi Xiong, Yongqi Zhang, Jiang Bian, Tianhao Peng, Xuhong Li, Shuaiqiang Wang, Linghe Kong, and Dawei Yin.
\newblock Rankelectra: Semi-supervised pre-training of learning-to-rank electra for web-scale search.
\newblock In {\em Proceedings of the 31st ACM SIGKDD Conference on Knowledge Discovery and Data Mining V. 1}, pages 2415--2425, 2025.

\bibitem{li2025m2oerank}
Yuchen Li, Hao Zhang, Yongqi Zhang, Xinyu Ma, Wenwen Ye, Naifei Song, Shuaiqiang Wang, Haoyi Xiong, Dawei Yin, and Lei Chen.
\newblock M2oerank: Multi-objective mixture-of-experts enhanced ranking for satisfaction-oriented web search.
\newblock In {\em 2025 IEEE 41st International Conference on Data Engineering (ICDE)}, pages 4441--4454. IEEE Computer Society, 2025.

\bibitem{li2025towards}
Yuchen Li, Hengyi Cai, Rui Kong, Xinran Chen, Jiamin Chen, Jun Yang, Haojie Zhang, Jiayi Li, Jiayi Wu, Yiqun Chen, et~al.
\newblock Towards ai search paradigm.
\newblock {\em arXiv preprint arXiv:2506.17188}, 2025.

\bibitem{li2023s2phere}
Yuchen Li, Haoyi Xiong, Linghe Kong, Qingzhong Wang, Shuaiqiang Wang, Guihai Chen, and Dawei Yin.
\newblock S2phere: Semi-supervised pre-training for web search over heterogeneous learning to rank data.
\newblock In {\em Proceedings of the 29th ACM SIGKDD Conference on Knowledge Discovery and Data Mining}, pages 4437--4448, 2023.

\bibitem{li2025rankexpert}
Yuchen Li, Hao Zhang, Yongqi Zhang, Hengyi Cai, Mingxin Cai, Shuaiqiang Wang, Haoyi Xiong, Dawei Yin, and Lei Chen.
\newblock Rankexpert: A mixture of textual-and-behavioral experts for multi-objective learning-to-rank in web search.
\newblock In {\em Proceedings of the 31st ACM SIGKDD Conference on Knowledge Discovery and Data Mining V. 2}, pages 4437--4449, 2025.

\bibitem{li2023coltr}
Yuchen Li, Haoyi Xiong, Qingzhong Wang, Linghe Kong, Hao Liu, Haifang Li, Jiang Bian, Shuaiqiang Wang, Guihai Chen, Dejing Dou, et~al.
\newblock Coltr: Semi-supervised learning to rank with co-training and over-parameterization for web search.
\newblock {\em IEEE Transactions on Knowledge and Data Engineering}, 35(12):12542--12555, 2023.

\bibitem{li2023mhrr}
Yuchen Li, Haoyi Xiong, Linghe Kong, Rui Zhang, Fanqin Xu, Guihai Chen, and Minglu Li.
\newblock Mhrr: Moocs recommender service with meta hierarchical reinforced ranking.
\newblock {\em IEEE Transactions on Services Computing}, 16(6):4467--4480, 2023.

\bibitem{li2025fultr}
Yuchen Li, Hao Zhang, Haojie Zhang, Hengyi Cai, Xinyu Ma, Shuaiqiang Wang, Haoyi Xiong, Zhaochun Ren, Maarten de~Rijke, and Dawei Yin.
\newblock Fultr: A large-scale fusion learning to rank dataset and its application for satisfaction-oriented ranking.
\newblock In {\em Proceedings of the 31st ACM SIGKDD Conference on Knowledge Discovery and Data Mining V. 2}, pages 5583--5594, 2025.

\bibitem{yu2025rainy}
Zhenyu Yu, Hanqing Chen, Mohd Yamani~Idna Idris, and Pei Wang.
\newblock Rainy: Unlocking satellite calibration for deep learning in precipitation.
\newblock {\em arXiv preprint arXiv:2504.10776}, 2025.

\bibitem{yu2025satellitemaker}
Zhenyu Yu, Mohd Yamani~Inda Idris, and Pei Wang.
\newblock Satellitemaker: A diffusion-based framework for terrain-aware remote sensing image reconstruction.
\newblock {\em arXiv preprint arXiv:2504.12112}, 2025.

\bibitem{yu2025forgetme}
Zhenyu Yu, Mohd Yamani~Inda Idris, and Pei Wang.
\newblock Forgetme: Evaluating selective forgetting in generative models.
\newblock {\em arXiv preprint arXiv:2504.12574}, 2025.

\bibitem{yu2025satellitecalculator}
Zhenyu Yu, Mohd Idris, and Pei Wang.
\newblock Satellitecalculator: A multi-task vision foundation model for quantitative remote sensing inversion.
\newblock {\em arXiv preprint arXiv:2504.13442}, 2025.

\bibitem{yu2025dancetext}
Zhenyu Yu, Mohd Yamani~Idna Idris, Pei Wang, and Yuelong Xia.
\newblock Dancetext: Point-driven interactive text and image layer editing using diffusion models.
\newblock {\em arXiv preprint arXiv:2504.14108}, 2025.

\bibitem{yu2025dc4cr}
Zhenyu Yu, Mohd Yamani~Idna Idris, and Pei Wang.
\newblock Dc4cr: When cloud removal meets diffusion control in remote sensing.
\newblock {\em arXiv preprint arXiv:2504.14785}, 2025.

\bibitem{yu2025satelliteformula}
Zhenyu Yu, Mohd Idris, Pei Wang, Yuelong Xia, Fei Ma, Rizwan Qureshi, et~al.
\newblock Satelliteformula: Multi-modal symbolic regression from remote sensing imagery for physics discovery.
\newblock {\em arXiv preprint arXiv:2506.06176}, 2025.

\bibitem{yu2025physics}
Zhenyu Yu, Mohd Yamani~Idna Idris, Hua Wang, Pei Wang, Junyi Chen, and Kun Wang.
\newblock From physics to foundation models: A review of ai-driven quantitative remote sensing inversion.
\newblock {\em arXiv preprint arXiv:2507.09081}, 2025.

\bibitem{ren2025estimating}
Jintong Ren, Lizhi Liu, You Wu, Lijian Ouyang, and Zhenyu Yu.
\newblock Estimating forest carbon stock using enhanced resnet and sentinel-2 imagery.
\newblock {\em Forests (19994907)}, 16(7), 2025.

\bibitem{sarkar2025reasoning}
Ayushman Sarkar, Mohd Yamani~Idna Idris, and Zhenyu Yu.
\newblock Reasoning in computer vision: Taxonomy, models, tasks, and methodologies.
\newblock {\em arXiv preprint arXiv:2508.10523}, 2025.

\bibitem{Diao_2025_WACV}
Xingjian Diao, Ming Cheng, Wayner Barrios, and SouYoung Jin.
\newblock Ft2tf: First-person statement text-to-talking face generation.
\newblock In {\em Proceedings of the Winter Conference on Applications of Computer Vision (WACV)}, pages 4821--4830, February 2025.

\bibitem{diao2025temporal}
Xingjian Diao, Chunhui Zhang, Weiyi Wu, Zhongyu Ouyang, Peijun Qing, Ming Cheng, Soroush Vosoughi, and Jiang Gui.
\newblock Temporal working memory: Query-guided segment refinement for enhanced multimodal understanding.
\newblock {\em arXiv preprint arXiv:2502.06020}, 2025.

\bibitem{diao2025learning}
Xingjian Diao, Tianzhen Yang, Chunhui Zhang, Weiyi Wu, Ming Cheng, and Jiang Gui.
\newblock Learning sparsity for effective and efficient music performance question answering.
\newblock {\em arXiv preprint arXiv:2506.01319}, 2025.

\bibitem{diao-etal-2024-learning}
Xingjian Diao, Chunhui Zhang, Tingxuan Wu, Ming Cheng, Zhongyu Ouyang, Weiyi Wu, and Jiang Gui.
\newblock Learning musical representations for music performance question answering.
\newblock In {\em Findings of the Association for Computational Linguistics: EMNLP 2024}, 2024.

\bibitem{diao2025soundmind}
Xingjian Diao, Chunhui Zhang, Keyi Kong, Weiyi Wu, Chiyu Ma, Zhongyu Ouyang, Peijun Qing, Soroush Vosoughi, and Jiang Gui.
\newblock Soundmind: Rl-incentivized logic reasoning for audio-language models.
\newblock {\em arXiv preprint arXiv:2506.12935}, 2025.

\bibitem{encoder}
Zixu Li, Zhiwei Chen, Haokun Wen, Zhiheng Fu, Yupeng Hu, and Weili Guan.
\newblock Encoder: Entity mining and modification relation binding for composed image retrieval.
\newblock In {\em Proceedings of the AAAI Conference on Artificial Intelligence}, volume~39, pages 5101--5109, 2025.

\bibitem{FineCIR}
Zixu Li, Zhiheng Fu, Yupeng Hu, Zhiwei Chen, Haokun Wen, and Liqiang Nie.
\newblock Finecir: Explicit parsing of fine-grained modification semantics for composed image retrieval.
\newblock {\em https://arxiv.org/abs/2503.21309}, 2025.

\bibitem{offset}
Zhiwei Chen, Yupeng Hu, Zixu Li, Zhiheng Fu, Xuemeng Song, and Liqiang Nie.
\newblock Offset: Segmentation-based focus shift revision for composed image retrieval, 2025.

\bibitem{MEDIAN}
Qinlei Huang, Zhiwei Chen, Zixu Li, Chunxiao Wang, Xuemeng Song, Yupeng Hu, and Liqiang Nie.
\newblock Median: Adaptive intermediate-grained aggregation network for composed image retrieval.
\newblock In {\em Proceedings of the IEEE International Conference on Acoustics, Speech and Signal Processing}, pages 1--5. IEEE, 2025.

\bibitem{PAIR}
Zhiheng Fu, Zixu Li, Zhiwei Chen, Chunxiao Wang, Xuemeng Song, Yupeng Hu, and Liqiang Nie.
\newblock Pair: Complementarity-guided disentanglement for composed image retrieval.
\newblock In {\em Proceedings of the IEEE International Conference on Acoustics, Speech and Signal Processing}, pages 1--5. IEEE, 2025.

\bibitem{qiu2024tfb}
Xiangfei Qiu, Jilin Hu, Lekui Zhou, Xingjian Wu, Junyang Du, Buang Zhang, Chenjuan Guo, Aoying Zhou, Christian~S. Jensen, Zhenli Sheng, and Bin Yang.
\newblock {TFB}: Towards comprehensive and fair benchmarking of time series forecasting methods.
\newblock In {\em Proc. {VLDB} Endow.}, pages 2363--2377, 2024.

\bibitem{qiu2025duet}
Xiangfei Qiu, Xingjian Wu, Yan Lin, Chenjuan Guo, Jilin Hu, and Bin Yang.
\newblock {DUET}: Dual clustering enhanced multivariate time series forecasting.
\newblock In {\em SIGKDD}, pages 1185--1196, 2025.

\bibitem{qiu2025tab}
Xiangfei Qiu, Zhe Li, Wanghui Qiu, Shiyan Hu, Lekui Zhou, Xingjian Wu, Zhengyu Li, Chenjuan Guo, Aoying Zhou, Zhenli Sheng, Jilin Hu, Christian~S. Jensen, and Bin Yang.
\newblock Tab: Unified benchmarking of time series anomaly detection methods.
\newblock In {\em Proc. {VLDB} Endow.}, pages 2775--2789, 2025.

\bibitem{chen2023rgp}
Zhuangzhi Chen, Jingyang Xiang, Yao Lu, Qi~Xuan, Zhen Wang, Guanrong Chen, and Xiaoniu Yang.
\newblock Rgp: Neural network pruning through regular graph with edges swapping.
\newblock {\em IEEE Transactions on Neural Networks and Learning Systems}, 35(10):14671--14683, 2023.

\bibitem{lu2022understanding}
Yao Lu, Wen Yang, Yunzhe Zhang, Zuohui Chen, Jinyin Chen, Qi~Xuan, Zhen Wang, and Xiaoniu Yang.
\newblock Understanding the dynamics of dnns using graph modularity.
\newblock In {\em European Conference on Computer Vision}, pages 225--242. Springer, 2022.

\bibitem{lu2024generic}
Yao Lu, Yutao Zhu, Yuqi Li, Dongwei Xu, Yun Lin, Qi~Xuan, and Xiaoniu Yang.
\newblock A generic layer pruning method for signal modulation recognition deep learning models.
\newblock {\em IEEE Transactions on Cognitive Communications and Networking}, 2024.

\bibitem{lu2024reassessing}
Yao Lu, Hao Cheng, Yujie Fang, Zeyu Wang, Jiaheng Wei, Dongwei Xu, Qi~Xuan, Xiaoniu Yang, and Zhaowei Zhu.
\newblock Reassessing layer pruning in llms: New insights and methods.
\newblock {\em arXiv preprint arXiv:2411.15558}, 2024.

\bibitem{lu2023can}
Yao Lu, Xuguang Chen, Yuchen Zhang, Jianyang Gu, Tianle Zhang, Yifan Zhang, Xiaoniu Yang, Qi~Xuan, Kai Wang, and Yang You.
\newblock Can pre-trained models assist in dataset distillation?
\newblock {\em arXiv preprint arXiv:2310.03295}, 2023.

\bibitem{lu2025llm}
Yao Lu, Zhaiyuan Ji, Jiawei Du, Yu~Shanqing, Qi~Xuan, and Tianyi Zhou.
\newblock From llm-anation to llm-orchestrator: Coordinating small models for data labeling.
\newblock {\em arXiv preprint arXiv:2506.16393}, 2025.

\bibitem{lu2024redtest}
Yao Lu, Peixin Zhang, Jingyi Wang, Lei Ma, Xiaoniu Yang, and Qi~Xuan.
\newblock Redtest: Towards measuring redundancy in deep neural networks effectively.
\newblock {\em arXiv preprint arXiv:2411.10507}, 2024.

\bibitem{li2024sglp}
Yuqi Li, Yao Lu, Zeyu Dong, Chuanguang Yang, Yihao Chen, and Jianping Gou.
\newblock Sglp: A similarity guided fast layer partition pruning for compressing large deep models.
\newblock {\em arXiv preprint arXiv:2410.14720}, 2024.

\bibitem{li2025sepprune}
Yuqi Li, Kai Li, Xin Yin, Zhifei Yang, Junhao Dong, Zeyu Dong, Chuanguang Yang, Yingli Tian, and Yao Lu.
\newblock Sepprune: Structured pruning for efficient deep speech separation.
\newblock {\em arXiv preprint arXiv:2505.12079}, 2025.

\bibitem{chen2021graph}
Zuohui Chen, Yao Lu, JinXuan Hu, Wen Yang, Qi~Xuan, Zhen Wang, and Xiaoniu Yang.
\newblock Graph-based similarity of neural network representations.
\newblock {\em arXiv preprint arXiv:2111.11165}, 2021.

\bibitem{lu2025fcos}
Yao Lu, Tengfei Ma, Zeyu Wang, Zhuangzhi Chen, Dongwei Xu, Yun Lin, Qi~Xuan, and Guan Gui.
\newblock Fcos: A two-stage recoverable model pruning framework for automatic modulation recognition.
\newblock {\em arXiv preprint arXiv:2505.21571}, 2025.

\bibitem{lu2025duse}
Yao Lu, Hongyu Gao, Zhuangzhi Chen, Dongwei Xu, Yun Lin, Qi~Xuan, and Guan Gui.
\newblock Duse: A data expansion framework for low-resource automatic modulation recognition based on active learning.
\newblock {\em arXiv preprint arXiv:2507.12011}, 2025.

\bibitem{tang2023sr}
Hui Tang, Yao Lu, and Qi~Xuan.
\newblock Sr-init: An interpretable layer pruning method.
\newblock In {\em ICASSP 2023-2023 IEEE International Conference on Acoustics, Speech and Signal Processing (ICASSP)}, pages 1--5. IEEE, 2023.

\bibitem{zhang2025frect}
Wenxin Zhang, Ding Xu, Guangzhen Yao, Xiaojian Lin, Renxiang Guan, Chengze Du, Renda Han, Xi~Xuan, and Cuicui Luo.
\newblock Frect: Frequency-augmented convolutional transformer for robust time series anomaly detection.
\newblock In {\em International Conference on Intelligent Computing}, pages 15--26. Springer, 2025.

\bibitem{shen2025mess}
Yaomin Shen, Xiaojian Lin, and Wei Fan.
\newblock A-mess: Anchor based multimodal embedding with semantic synchronization for multimodal intent recognition.
\newblock {\em arXiv preprint arXiv:2503.19474}, 2025.

\bibitem{zhang2025dual}
Wenxin Zhang, Jingxing Zhong, Guangzhen Yao, Renda Han, Xiaojian Lin, Zeyu Zhang, and Cuicui Luo.
\newblock Dual-channel heterophilic message passing for graph fraud detection.
\newblock {\em arXiv preprint arXiv:2504.14205}, 2025.

\bibitem{zhang2025dconad}
Wenxin Zhang, Xiaojian Lin, Wenjun Yu, Guangzhen Yao, Yu~Li, Renda Han, Songcheng Xu, Hao Shi, Cuicui Luo, et~al.
\newblock Dconad: A differencing-based contrastive representation learning framework for time series anomaly detection.
\newblock {\em arXiv preprint arXiv:2504.14204}, 2025.

\bibitem{price2018combining}
Nicholas Price, Brook~T Moyers, Lua Lopez, Jesse~R Lasky, J~Grey Monroe, Jack~L Mullen, Christopher~G Oakley, Junjiang Lin, Jon {\AA}gren, Daniel~R Schrider, et~al.
\newblock Combining population genomics and fitness qtls to identify the genetics of local adaptation in arabidopsis thaliana.
\newblock {\em Proceedings of the National Academy of Sciences}, 115(19):5028--5033, 2018.

\bibitem{fletcher2016identification}
Richard~S Fletcher, David Herrmann, Jack~L Mullen, Qinfei Li, Daniel~R Schrider, Nicholas Price, Junjiang Lin, Kelsi Grogan, Andrew Kern, and John~K McKay.
\newblock Identification of polymorphisms associated with drought adaptation qtl in brassica napus by resequencing.
\newblock {\em G3: Genes, Genomes, Genetics}, 6(4):793--803, 2016.

\bibitem{price2020linking}
Nicholas Price, Jack~L Mullen, Junjiang Lin, Christina Boucher, and John~K McKay.
\newblock Linking genomic signatures of selection to expression variation and direct evidence of local adaptation.
\newblock {\em bioRxiv}, pages 2020--08, 2020.

\bibitem{stupp2005radiotherapy}
Roger Stupp, Warren~P Mason, Martin~J Van Den~Bent, Michael Weller, Barbara Fisher, Martin~JB Taphoorn, Karl Belanger, Alba~A Brandes, Christine Marosi, Ulrich Bogdahn, et~al.
\newblock Radiotherapy plus concomitant and adjuvant temozolomide for glioblastoma.
\newblock {\em New England journal of medicine}, 352(10):987--996, 2005.

\bibitem{davis2016glioblastoma}
Mary~Elizabeth Davis.
\newblock Glioblastoma: overview of disease and treatment.
\newblock {\em Clinical journal of oncology nursing}, 20(5):S2, 2016.

\bibitem{aldape2019challenges}
Kenneth Aldape, Kevin~M Brindle, Louis Chesler, Rajesh Chopra, Amar Gajjar, Mark~R Gilbert, Nicholas Gottardo, David~H Gutmann, Darren Hargrave, Eric~C Holland, et~al.
\newblock Challenges to curing primary brain tumours.
\newblock {\em Nature reviews Clinical oncology}, 16(8):509--520, 2019.

\bibitem{kickingereder2016radiomic}
Philipp Kickingereder, Sina Burth, Antje Wick, Michael G{\"o}tz, Oliver Eidel, Heinz-Peter Schlemmer, Klaus~H Maier-Hein, Wolfgang Wick, Martin Bendszus, Alexander Radbruch, et~al.
\newblock Radiomic profiling of glioblastoma: identifying an imaging predictor of patient survival with improved performance over established clinical and radiologic risk models.
\newblock {\em Radiology}, 280(3):880--889, 2016.

\bibitem{liang2024cmat}
Xuechen Liang, Yangfan He, Meiling Tao, Yinghui Xia, Jianhui Wang, Tianyu Shi, Jun Wang, and JingSong Yang.
\newblock Cmat: A multi-agent collaboration tuning framework for enhancing small language models.
\newblock {\em arXiv preprint arXiv:2404.01663}, 2024.

\bibitem{wang2024enhancing}
Junqiao Wang, Zeng Zhang, Yangfan He, Zihao Zhang, Yuyang Song, Tianyu Shi, Yuchen Li, Hengyuan Xu, Kunyu Wu, Xin Yi, et~al.
\newblock Enhancing code llms with reinforcement learning in code generation: A survey.
\newblock {\em arXiv preprint arXiv:2412.20367}, 2024.

\bibitem{zhou2024human}
Ziqi Zhou, Jingyue Zhang, Jingyuan Zhang, Yangfan He, Boyue Wang, Tianyu Shi, and Alaa Khamis.
\newblock Human-centric reward optimization for reinforcement learning-based automated driving using large language models.
\newblock {\em arXiv preprint arXiv:2405.04135}, 2024.

\bibitem{zhou2025reagent}
Yiyang Zhou, Yangfan He, Yaofeng Su, Siwei Han, Joel Jang, Gedas Bertasius, Mohit Bansal, and Huaxiu Yao.
\newblock Reagent-v: A reward-driven multi-agent framework for video understanding.
\newblock {\em arXiv preprint arXiv:2506.01300}, 2025.

\bibitem{yi2025score}
Qiang Yi, Yangfan He, Jianhui Wang, Xinyuan Song, Shiyao Qian, Xinhang Yuan, Li~Sun, Yi~Xin, Jingqun Tang, Keqin Li, et~al.
\newblock Score: Story coherence and retrieval enhancement for ai narratives.
\newblock {\em arXiv preprint arXiv:2503.23512}, 2025.

\bibitem{ellingson2015consensus}
Benjamin~M Ellingson, Martin Bendszus, Jerrold Boxerman, Daniel Barboriak, Bradley~J Erickson, Marion Smits, Sarah~J Nelson, Elizabeth Gerstner, Brian Alexander, Gregory Goldmacher, et~al.
\newblock Consensus recommendations for a standardized brain tumor imaging protocol in clinical trials.
\newblock {\em Neuro-oncology}, 17(9):1188--1198, 2015.

\bibitem{louis20212021}
David~N Louis, Arie Perry, Pieter Wesseling, Daniel~J Brat, Ian~A Cree, Dominique Figarella-Branger, Cynthia Hawkins, HK~Ng, Stefan~M Pfister, Guido Reifenberger, et~al.
\newblock The 2021 who classification of tumors of the central nervous system: a summary.
\newblock {\em Neuro-oncology}, 23(8):1231--1251, 2021.

\bibitem{bakas2018identifying}
Spyridon Bakas, Mauricio Reyes, Andras Jakab, Stefan Bauer, Markus Rempfler, Alessandro Crimi, Russell~Takeshi Shinohara, Christoph Berger, Sung~Min Ha, Martin Rozycki, et~al.
\newblock Identifying the best machine learning algorithms for brain tumor segmentation, progression assessment, and overall survival prediction in the brats challenge.
\newblock {\em arXiv preprint arXiv:1811.02629}, 2018.

\bibitem{miotto2018deep}
Riccardo Miotto, Fei Wang, Shuang Wang, Xiaoqian Jiang, and Joel~T Dudley.
\newblock Deep learning for healthcare: review, opportunities and challenges.
\newblock {\em Briefings in bioinformatics}, 19(6):1236--1246, 2018.

\bibitem{topol2019high}
Eric~J Topol.
\newblock High-performance medicine: the convergence of human and artificial intelligence.
\newblock {\em Nature medicine}, 25(1):44--56, 2019.

\bibitem{liang2024self}
Xuechen Liang, Yangfan He, Yinghui Xia, Xinyuan Song, Jianhui Wang, Meiling Tao, Li~Sun, Xinhang Yuan, Jiayi Su, Keqin Li, et~al.
\newblock Self-evolving agents with reflective and memory-augmented abilities.
\newblock {\em arXiv preprint arXiv:2409.00872}, 2024.

\bibitem{xin2025resurrect}
Yi~Xin, Le~Zhuo, Qi~Qin, Siqi Luo, Yuewen Cao, Bin Fu, Yangfan He, Hongsheng Li, Guangtao Zhai, Xiaohong Liu, et~al.
\newblock Resurrect mask autoregressive modeling for efficient and scalable image generation.
\newblock {\em arXiv preprint arXiv:2507.13032}, 2025.

\bibitem{zhou2025glimpse}
Yiyang Zhou, Linjie Li, Shi Qiu, Zhengyuan Yang, Yuyang Zhao, Siwei Han, Yangfan He, Kangqi Li, Haonian Ji, Zihao Zhao, et~al.
\newblock Glimpse: Do large vision-language models truly think with videos or just glimpse at them?
\newblock {\em arXiv preprint arXiv:2507.09491}, 2025.

\bibitem{he2025enhancing}
Yangfan He, Sida Li, Jianhui Wang, Kun Li, Xinyuan Song, Xinhang Yuan, Keqin Li, Kuan Lu, Menghao Huo, Jingqun Tang, et~al.
\newblock Enhancing low-cost video editing with lightweight adaptors and temporal-aware inversion.
\newblock {\em arXiv preprint arXiv:2501.04606}, 2025.

\bibitem{huo2025ct}
Menghao Huo, Kuan Lu, Yuxiao Li, Qiang Zhu, and Zhenrui Chen.
\newblock Ct-patchtst: Channel-time patch time-series transformer for long-term renewable energy forecasting.
\newblock {\em arXiv preprint arXiv:2501.08620}, 2025.

\bibitem{wang2025twin}
Jianhui Wang, Yangfan He, Yan Zhong, Xinyuan Song, Jiayi Su, Yuheng Feng, Hongyang He, Wenyu Zhu, Xinhang Yuan, Kuan Lu, et~al.
\newblock Twin co-adaptive dialogue for progressive image generation.
\newblock {\em arXiv preprint arXiv:2504.14868}, 2025.

\bibitem{zhong2025comparative}
Jiachen Zhong and Yiting Wang.
\newblock A comparative study of ensemble models for thyroid disease prediction under class imbalance.
\newblock 2025.

\bibitem{yang2025oral}
Runhua Yang, Hongyu Jin, Chenyu Zhao, Wei Wang, and Wen-Yang Li.
\newblock Oral cancer and sleep disturbances: A narrative review on exploring the bidirectional relationship.
\newblock {\em Cancers}, 17(8):1262, 2025.

\bibitem{wang2025applications}
Yiting Wang, Ziwei Wang, Jiachen Zhong, Di~Zhu, and Weiyi Li.
\newblock Applications of small language models in medical imaging classification with a focus on prompt strategies.
\newblock {\em arXiv preprint arXiv:2508.13378}, 2025.

\bibitem{zhao2023advances}
Chenyu Zhao, Boyue Pan, Tianlin Wang, Huazhe Yang, David Vance, Xiaojia Li, Haiyang Zhao, Xinru Hu, Tianchang Yang, Zihao Chen, et~al.
\newblock Advances in nir-responsive natural macromolecular hydrogel assembly drugs for cancer treatment.
\newblock {\em Pharmaceutics}, 15(12):2729, 2023.

\bibitem{zhao2024antibody}
Chenyu Zhao, Ruihan Zhang, Huazhe Yang, Yiwei Gao, Ying Zou, and Xudong Zhang.
\newblock Antibody-drug conjugates for non-small cell lung cancer: Advantages and challenges in clinical translation.
\newblock {\em Biochemical Pharmacology}, 226:116378, 2024.

\bibitem{lao2017deep}
Jiangwei Lao, Yinsheng Chen, Zhi-Cheng Li, Qihua Li, Ji~Zhang, Jing Liu, and Guangtao Zhai.
\newblock A deep learning-based radiomics model for prediction of survival in glioblastoma multiforme.
\newblock {\em Scientific reports}, 7(1):10353, 2017.

\bibitem{chaddad2025radiomic}
Ahmad Chaddad, Pingyue Jia, Yan Hu, Yousef Katib, Reem Kateb, and Tareef~Sahal Daqqaq.
\newblock A radiomic model for gliomas grade and patient survival prediction.
\newblock {\em Bioengineering}, 12(5):450, 2025.

\bibitem{ren2023multimodality}
Jinfa Ren, Xiaoyang Zhai, Huijia Yin, Fengmei Zhou, Ying Hu, Kaiyu Wang, Ruifang Yan, and Dongming Han.
\newblock Multimodality mri radiomics based on machine learning for identifying true tumor recurrence and treatment-related effects in patients with postoperative glioma.
\newblock {\em Neurology and Therapy}, 12(5):1729--1743, 2023.

\bibitem{wang2025systematic}
Yiting Wang, Jiachen Zhong, and Rohan Kumar.
\newblock A systematic review of machine learning applications in infectious disease prediction, diagnosis, and outbreak forecasting.
\newblock 2025.

\bibitem{zadeh2020deepsurvnet}
Amin Zadeh~Shirazi, Eric Fornaciari, Narjes~Sadat Bagherian, Lisa~M Ebert, Barbara Koszyca, and Guillermo~A Gomez.
\newblock Deepsurvnet: deep survival convolutional network for brain cancer survival rate classification based on histopathological images.
\newblock {\em Medical \& biological engineering \& computing}, 58(5):1031--1045, 2020.

\bibitem{akkus2017deep}
Zeynettin Akkus, Alfiia Galimzianova, Assaf Hoogi, Daniel~L Rubin, and Bradley~J Erickson.
\newblock Deep learning for brain mri segmentation: state of the art and future directions.
\newblock {\em Journal of digital imaging}, 30(4):449--459, 2017.

\bibitem{zhu2025image}
Qiang Zhu, Kuan Lu, Menghao Huo, and Yuxiao Li.
\newblock Image-to-image translation with diffusion transformers and clip-based image conditioning.
\newblock {\em arXiv preprint arXiv:2505.16001}, 2025.

\bibitem{gomaa2024comprehensive}
Ahmed Gomaa, Yixing Huang, Amr Hagag, Charlotte Schmitter, Daniel H{\"o}fler, Thomas Weissmann, Katharina Breininger, Manuel Schmidt, Jenny Stritzelberger, Daniel Delev, et~al.
\newblock Comprehensive multimodal deep learning survival prediction enabled by a transformer architecture: A multicenter study in glioblastoma.
\newblock {\em Neuro-Oncology Advances}, 6(1):vdae122, 2024.

\bibitem{luckett2023predicting}
Patrick~H Luckett, Michael Olufawo, Bidhan Lamichhane, Ki~Yun Park, Donna Dierker, Gabriel~Trevino Verastegui, Peter Yang, Albert~H Kim, Milan~G Chheda, Abraham~Z Snyder, et~al.
\newblock Predicting survival in glioblastoma with multimodal neuroimaging and machine learning.
\newblock {\em Journal of Neuro-oncology}, 164(2):309--320, 2023.

\bibitem{mahootiha2025multimodal}
Maryamalsadat Mahootiha, Divyanshu Tak, Zezhong Ye, Anna Zapaishchykova, Jirapat Likitlersuang, Juan~Carlos Climent~Pardo, Aidan Boyd, Sridhar Vajapeyam, Rishi Chopra, Sanjay~P Prabhu, et~al.
\newblock Multimodal deep learning improves recurrence risk prediction in pediatric low-grade gliomas.
\newblock {\em Neuro-oncology}, 27(1):277--290, 2025.

\bibitem{taouli2004magnetic}
Bachir Taouli, Mariela Losada, Agnes Holland, and Glenn Krinsky.
\newblock Magnetic resonance imaging of hepatocellular carcinoma.
\newblock {\em Gastroenterology}, 127(5):S144--S152, 2004.

\bibitem{zwanenburg2020image}
Alex Zwanenburg, Martin Valli{\`e}res, Mahmoud~A Abdalah, Hugo~JWL Aerts, Vincent Andrearczyk, Aditya Apte, Saeed Ashrafinia, Spyridon Bakas, Roelof~J Beukinga, Ronald Boellaard, et~al.
\newblock The image biomarker standardization initiative: standardized quantitative radiomics for high-throughput image-based phenotyping.
\newblock {\em Radiology}, 295(2):328--338, 2020.

\bibitem{imamura2003risk}
Hiroshi Imamura, Yutaka Matsuyama, Eiji Tanaka, Takao Ohkubo, Kiyoshi Hasegawa, Shinichi Miyagawa, Yasuhiko Sugawara, Masami Minagawa, Tadatoshi Takayama, Seiji Kawasaki, et~al.
\newblock Risk factors contributing to early and late phase intrahepatic recurrence of hepatocellular carcinoma after hepatectomy.
\newblock {\em Journal of hepatology}, 38(2):200--207, 2003.

\bibitem{galle2018easl}
Peter~R Galle, Alejandro Forner, Josep~M Llovet, Vincenzo Mazzaferro, Fabio Piscaglia, Jean-Luc Raoul, Peter Schirmacher, and Val{\'e}rie Vilgrain.
\newblock Easl clinical practice guidelines: management of hepatocellular carcinoma.
\newblock {\em Journal of hepatology}, 69(1):182--236, 2018.

\bibitem{cox1972regression}
David~R Cox.
\newblock Regression models and life-tables.
\newblock {\em Journal of the Royal Statistical Society: Series B (Methodological)}, 34(2):187--202, 1972.

\bibitem{ishwaran2008random}
Hemant Ishwaran, Udaya~B Kogalur, Eugene~H Blackstone, and Michael~S Lauer.
\newblock Random survival forests.
\newblock 2008.

\bibitem{lee2019dynamic}
Changhee Lee, Jinsung Yoon, and Mihaela Van Der~Schaar.
\newblock Dynamic-deephit: A deep learning approach for dynamic survival analysis with competing risks based on longitudinal data.
\newblock {\em IEEE Transactions on Biomedical Engineering}, 67(1):122--133, 2019.

\bibitem{vickers2006decision}
Andrew~J Vickers and Elena~B Elkin.
\newblock Decision curve analysis: a novel method for evaluating prediction models.
\newblock {\em Medical Decision Making}, 26(6):565--574, 2006.

\end{thebibliography}

\end{document}